\documentclass[11pt,a4paper]{article}
\usepackage{times}
\usepackage{graphicx}
\usepackage{amssymb}
\usepackage{url,hyperref}
\usepackage[hyperref]{acl2020}
\usepackage{amsmath}
\usepackage{booktabs}
\usepackage{tabularx}
\usepackage{multirow,bigdelim}

\usepackage{algpseudocode}
\usepackage{varwidth}
\usepackage{microtype}

\aclfinalcopy 

\title{Data Augmentation for Mental Health Classification on Social Media}

\author{Gunjan Ansari \\
   JSS Academy of Technical\\ Education, Noida, India\\
  \texttt{gunjanansari@jssaten.ac.in} \\\And
  Muskan Garg\\
  Thapar Institute of \\Engineering \& Technology\\
  Patiala, Punjab,\\ India\\
  \texttt{muskanphd@gmail.com} \\ \And
  Chandni Saxena\\
  The Chinese University\\ of Hong Kong, Shatin,\\ NT, Hong Kong \\
  \texttt{csaxena@cse.cuhk.edu.hk} }

\date{}

\begin{document}
\maketitle
\begin{abstract}
The mental disorder of online users is determined using social media posts. The major challenge in this domain is to avail the ethical clearance for using the user-generated text on social media platforms. Academic researchers identified the problem of insufficient and unlabeled data for mental health classification. To handle this issue, we have studied the effect of data augmentation techniques on domain-specific user-generated text for mental health classification. Among the existing well-established data augmentation techniques, we have identified Easy Data Augmentation (EDA), conditional BERT, and Back-Translation (BT) as the potential techniques for generating additional text to improve the performance of classifiers. Further, three different classifiers- Random Forest (RF), Support Vector Machine (SVM) and Logistic Regression (LR) are employed for analyzing the impact of data augmentation on two publicly available social media datasets. The experimental results show significant improvements in classifiers' performance when trained on the augmented data.  
\end{abstract}

\section{Introduction}

Recent studies over mental health classification \cite{salari2020prevalence, garg2021quantifying, biester2021understanding} convey that amid COVID-19 pandemic, the number of stress, anxiety and depression related mental disorders have increased. As per the recent survey, the rate of increase of mental disorders is more than those of physical health impacts on the Chinese population~\cite{huang2020generalized}. In this context, the early detection of psychological disorders is very important for good governance. It is observed that more than 80\% of the people who commit suicide, disclose their intention to do so on social media~\cite{sawhney2021phase}. Clinical depression is the result of frequent tensions and stress. Further, prevailing clinical depression for a longer time period results in suicidal tendencies.   

The information mining from social media helps in identifying stressful and casual conversations \cite{thelwall2017tensistrength, Turcan2019DreadditAR, turcan2021emotion}. Many Machine Learning (ML) algorithms are developed in literature using both automatic and handcrafted features for classifying Microblog. The problem of data sparsity is underexplored for mental health studies on social media due to the sensitivity of data \cite{wongkoblap2017researching}. Multiple ethical clearances are required for new developments in mental health classification. To deal with this issue of data sparsity, we have used data augmentation techniques to multiply the training data \cite{Turcan2019DreadditAR, haque2021deep}. The increase in training data may help to improve the hyper-parameter learning of textual features and thereby, reducing overfitting. Data Augmentation is the method of increasing the data diversity without collecting more data \cite{feng2021survey}. The idea behind the use of Data Augmentation (DA) techniques is to understand the improvements in training classifiers for mental health detection on social media. 

In this manuscript, the mental health classification is performed for two datasets to test the scalability of data augmentation approaches for mental healthcare domain. The classification of casual and stressful conversations \cite{Turcan2019DreadditAR}, and classifying depression and suicidal posts \cite{haque2021deep} on social media. We select a rule based approach which preserves the original label and diversifies the text. To the best of our knowledge, this is the first attempt of stuffing additional data for mental health classification and there is no such study in the existing literature. The key contributions of this work are as follows:
\begin{itemize}
    \item To determine the feasibility and the importance of data augmentation in the domain-specific study of mental health classification to solve the problem of data sparsity.
    \item The empirical study for different classification algorithms show significantly improved F-measure. 
\end{itemize}

Ethical Clearance: We use limited, sparse and publicly available dataset for this study and so, no ethical approval is required from the Institutional Review Board (IRB) or elsewhere. 

We organize rest of the manuscript in different sections. Section~2 describes the historical perspective of data augmentation and mental health classification on social media. We discuss the data augmentation methods and the architecture for experimental setups in Section~3. Section~4 elucidates the experimental results and evaluation over the proposed architecture of experimental setup which shows the significance and feasibility of data augmentation over mental health classification problems. Finally, Section~5 gives the conclusion and future scope of this work.

\section{Related Work}
Mental health classification can be quite challenging without the availability of sufficient data. Although the  users' posts can be extracted from the social media platforms such as Reddit, Twitter and Facebook, annotating these posts is quite expensive. To address this issue, researchers have proposed different data augmentation techniques suitable for Natural Language Processing (NLP) which varies from simple rule-based methods to more complex generative approaches  \cite{feng2021survey}. The data augmentation tasks is categorized into conditional and unconditional augmentation task \cite{shorten2021text}. 

\subsection{Evolution of textual Data Augmentation}
The unconditional data augmentation models like \emph{Generative adversarial networks} \cite{goodfellow2014generative} and \emph{Variational autoencoders} \cite{kingma2014autoencoding} generates the random texts irrespective of the context. We do not use unconditional data augmentation for this task as it is required to preserve the context of the information as per the label. The conditional masking of a few tokens in the original sentence was observed to boost the classification performance in NLP tasks~\cite{li2020conditional, wu2021conditional}. Bidirectional Encoder Representations from
Transformers (BERT) \cite{devlin2019bert}, the pre-trained language models, are proposed with the objective to capture the left and right context in the sentence to generate the masked tokens. The pre-trained autoencoder model conditional BERT \cite{wu2019conditional, kumar2021data} is used as a well-established technique for generating label compatible augmented data from the original data.

One of the simplest rule-based data augmentation techniques is proposed as Easy Data Augmentation (EDA)~\cite{wei2019eda}. The authors proposed four random operations such as \textit{random insertion, random deletion, random swapping and random replacement} on the given text for generating new sentences. The experimental results give better performance on five benchmark text classification tasks \cite{wei2019eda}, as the true labels of the generated text were conserved during the process of \emph{data augmentation}. A graph based data augmentation is proposed for sentences using balance theory and transitivity to infer the pairs generated by augmentation of sentences \cite{chen2020finding}. The sentence-based data augmentation is not suitable for the problem of mental health classification on Reddit data as the posts contain large paragraphs. 

Back Translation (BT) or Round-trip translation is another augmentation technique which is used as a pipeline for text generation \cite{sennrich2015improving}. The BT approach converts the $A$ language of text to $B$ language of text and then back to $A$ language of the same text. This back-translation~\cite{corbeil2020bet} of data helps in diversifying the data by preserving its contextual information. Although, the interpolation techniques are proposed for data augmentation \cite{zhang2017mixup}, it is minimally used for textual data in existing literature \cite{guo2020sequence}.  

In our work, we have studied the effect of all three different augmentation techniques- EDA, Conditional BERT and Back-translation to increase the size of training data for the task of mental health classification.

\subsection{Mental Health Classification: Historical Perspective}

The existing literature on mental health detection and analysis of social media data \cite{garg2021quantifying} shows the problem of automatic labeling as \emph{noisy labels}. To handle this, either the label correction of noisy labels is required as shown in SDCNL~\cite{haque2021deep} for manual labeling, or data augmentation \cite{chen2021empirical}. Since many existing datasets for mental health detection like RSDD, SMHD \cite{harrigian2020models}, CLPsych \cite{preoctiuc2015mental} needs ethical clearance and are available only on request, we intend to pick small dataset with limited set of instances which are available in the public domain.

The Dreaddit dataset is manually labelled as stressful and casual conversation ~\cite{Turcan2019DreadditAR}. In SDCNL dataset \cite{haque2021deep}, the posts related to clinical depression and suicidal tendencies use similar words. Thus, we hypothesize that experimental results with data augmentation for classifying depression and suicidal risk may not generate well diversified data. In this manuscript, we use three data augmentation methods to text and validate the performance of the classifiers over both Dreaddit and SDCNL dataset.   

\section{Background: Data Augmentation Methods}
Data augmentation~\cite{feng2021survey} is a recent technique used for NLP to handle the problem of data sparsity by increasing the size of the training data without explicitly collecting the data. In this Section, we describe three potential textual data augmentation techniques, problem formulation, and architecture of the experimental setup.  

\subsection{Textual Data Augmentation}
Out of many data augmentation tasks for NLP classification, very few are related to this problem domain of mental healthcare. This limitation is due to the presence of ill-formed (user-generated) text and the need to preserve the contextual information as per the label of the instances. To handle this issue, we use three different approaches. The first approach is based on NLP-based Augmentation technique~\cite{wei2019eda}, the second is based on conditional pre-trained language models such as BERT ~\cite{kumar2021data} and the third approach is based on back translation~\cite{ng2019facebook}. We briefly explain these methods in this section.
\subsubsection{Easy Data Augmentation}
In the previous work~\cite{wei2019eda}, NLP-based operations have been shown to achieve good results on text classification tasks. This method of data augmentation helps in diversifying the training samples while maintaining the class label associated with the post of a user at sentence level. The following four operations have been used in this work for augmenting the data:  
\begin{itemize}

    \item \textbf{Synonym Replacement.} Randomly $n$-words are chosen other than stop words from each sentence and replaced by one of its synonyms. 
    \item \textbf{Random Insertion.} In this operation, a random synonym of a random word is inserted into a random position of a sentence for n number of times.
    \item \textbf{Random Swap.} Two words are randomly chosen in a sentence and swapped.
    \item \textbf{Random Deletion.} A word is deleted from a sentence with probability $p$.
\end{itemize}

\subsubsection{Pre-Trained Language Models}
Recently, deep bi-directional models have been used for generating textual data \cite{kobayashi2018contextual,song2019mass,dong2017learning}. These models are pre-trained with unlabelled text which can be fine tuned in autoencoder ~\cite{devlin2019bert}, auto-regressive ~\cite{radford2019language}, or seq2seq~\cite{lewis2019bart} settings. In \emph{autoencoder settings}, a few tokens are randomly masked and the model is trained to predict alternative tokens. In \emph{auto-regressive settings}, the model predicts the succeeding word according to the context. In \emph{seq2seq settings}, the model is fine tuned on denoising autoencoder tasks. These transformers use associated class labels to generate the augmented text which helps in preserving its label. In this work, we adopt a framework\footnote{https://github.com/amazon-research/transformers-data-augmentation} defined by~\cite{kumar2021data} and fine tune  pre-trained BERT in auto-regressive settings.



\subsubsection{Back Translation}
Back translation (BT) is the data augmentation technique used for diversifying the information by changing the language of textual data to some language A and changing it back to its original language. In this experimental framework, we have used German as an intermediate language A. We use BT for the Microblogs by first converting it into German language using Neural Machine Translation \cite{ng2019facebook} and then converting it back to the English language. It is interesting to note that ill-formed and user-generated information is converted to the standard English language using BT and thus, spelling mistakes are reduced. Although the content is changed, contextual information is preserved. 

\subsection{Problem Formulation}
Given a dataset $D$ consisting of n-training samples where each sample is a text sequence $x$ consisting of $m$-words and each sequence is associated with a label $y$. The objective is to generate an augmented data D\textsubscript{syn} of n-synthetic samples using EDA, BERT and Back Translation. 

\subsubsection{AugEDA: Data augmentation using Easy Data Augmentation }
In our work, 30\% words of i\textsuperscript{th} training sample are randomly chosen for applying any one of the four EDA operation-Synonym Replacement, Random Insertion, Random Swap and Random Deletion \cite{wei2019eda}. In synonym replacement, the chosen word is substituted by any one of the randomly selected synonym of this word from WordNet~\cite{miller1995wordnet}. In random insertion, j random positions are chosen for inserting random synonym of randomly chosen word out of m-words. In random swap, two words are randomly chosen from m-words and swapped with each other. A word is deleted with 10\% probability in random deletion operation. The new sentence generated after applying any one of the lexical substitution method is added to the synthetic dataset D\textsubscript{syn}. The process is repeated for n-training samples to create an augmented dataset of size $n$.

\subsubsection{AugBERT: Data augmentation using BERT}
We use the conditional BERT language model to generate the augmented data. We consider the label $y$ and sequence $ S={S_1,S_2...S_N} $ of n-tokens to calculate the probability  $p (t_i) =(.|y,S) $ of masked token $ {t_i} $ unlike masked language models that use only sequence $ S $ for predicting the probability of masked tokens. As defined by~\cite{kumar2021data}, the conditional BERT model prepends associated label $y$ to each sequence $ S $ in dataset $D$ without adding it to the vocabulary of the model. For fine tuning of the model, some tokens of the sequence are randomly masked and the objective is to predict the original token according to the context of the sequence. 
\subsubsection{AugBT: Data augmentation using Back-Translation}
To generate new textual data using Back-Translation, each of  i\textsuperscript{th} training sample ${x_i}$  is converted into a sentence ${y_i}$ written in German language and then  ${y_i}$ is converted back to a sentence ${z_i}$ in English.  The generated sentence ${z_i}$ is added to the augmented dataset D\textsubscript{syn}. This process is repeated for $n$ training samples to create an augmented dataset of $n$ samples.

\subsection{Architecture: Experimental Setup}
\begin{figure}
    \begin{center}
    
    \includegraphics[width=80mm]{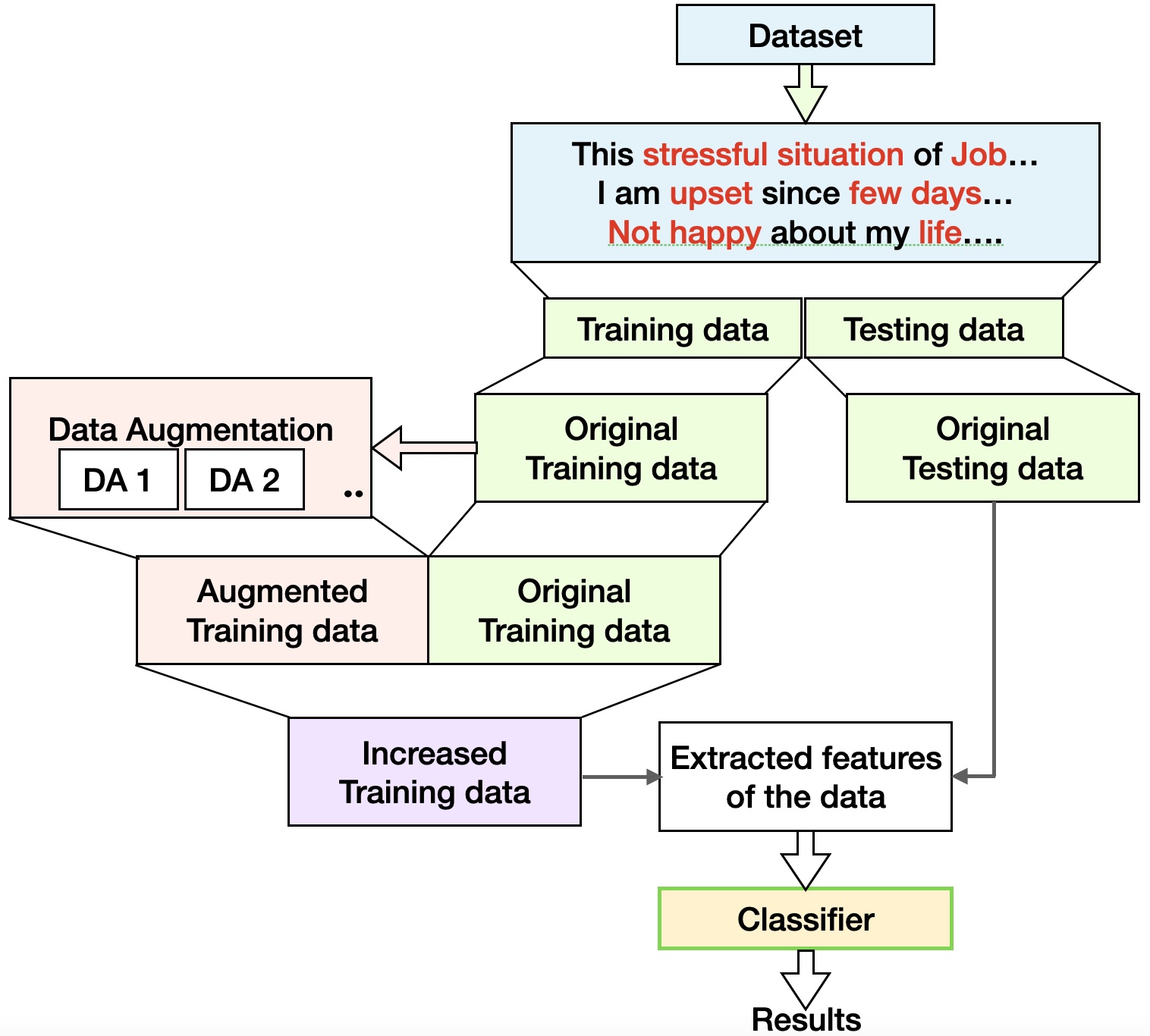}
    \caption{The Architecture of Experimental Setup for Data Augmentation}
   \label{fig:architecture} 
       
    \end{center}
\end{figure}

The architecture of the experimental setup for augmenting domain-specific data of mental health classification from social media posts is shown in Figure~\ref{fig:architecture}. The Microblogs are given as an input for classifying the mental health of the users. The idea behind this approach is to generate some sequence of sentences and augment some more data for better training of classifiers. Thus, the number of instances are increased by using different data augmentation techniques. 

The results are implemented for two publicly available mental health datasets, namely, Dreaddit and SDCNL. The dataset is divided into training and testing data. The training data is given as an input to the data augmentation methodologies, namely, EDA \cite{wei2019eda}, Autoencoder conditional BERT \cite{wu2019conditional} and Back-Translation \cite{ng2019facebook}. These three approaches are well established approaches for data augmentation in classification of the textual data. The original training data is almost doubled in the process of the data augmentation. The original and augmented data are fed to different machine learning classifiers for results and analysis. 

\section{Experimental Results and Evaluation}
In this section, we discuss the datasets and the experimental results. We further analyze results for data diversity and statistical significance of the classifiers over augmented data as compared to the original data.

\subsection{Dataset}
The idea behind this study is to improve the training parameters of the classifier by removing the limitation of data sparsity. The two sparse datasets which are used for domain-specific data augmentation are Dreaddit \footnote{http:
//www.cs.columbia.edu/˜eturcan/data/
dreaddit.zip.} \cite{Turcan2019DreadditAR} and SDCNL\footnote{https://github.com/ayaanzhaque/SDCNL} \cite{haque2021deep} from existing literature are explained in this Section.

\subsubsection{Dreaddit dataset}
The Dreaddit dataset\cite{Turcan2019DreadditAR} consists of lengthy posts in five different categories and is used for classifying stressful posts from casual conversations. The categories of subreddits selected by authors having stressful conversations are interpersonal conflicts, mental illness (anxiety and PTSD), financial and social. 

\begin{table}[h!]
  \begin{center}
\label{tab:table1}
    \begin{tabular}{l|c|c} 
      \textbf{Dataset } & \textbf{Stress} & \textbf{Non-Stress} \\
      
      \hline
      Training data  & 1488 & 1350 \\
      Testing data  & 369 & 346 \\
    \end{tabular}
  \end{center}
  \caption{Dreaddit Dataset Statistics}
\end{table}

Out of total $187444$ posts scraped from these five categories, the authors have manually labelled $3553$ Reddit posts. While selecting the posts for annotation, the authors selected those segments whose average token length was greater than $100$. The average tokens per post in this dataset is $420$ tokens. This statistics of the Dreaddit dataset is shown in Table~1. 

\subsubsection{SDCNL dataset}
The SDCNL dataset\cite{haque2021deep} is scrapped from Reddit social media platform from two subreddits: \emph{r/SuicideWatch} and \emph{r/Depression} to carry out the study for classifying posts into depression specific or suicide specific. This dataset contains 1895 posts containing $1517$ training samples and $379$ testing samples. The dataset contains title, selftext and megatext of the reddit tweets along with other fields. 

\begin{table}[h!]
  \begin{center}
    
    \label{tab:table2}
    \begin{tabular}{l|c|c} 
      \textbf{Dataset } & \textbf{Depression} & \textbf{Suicide} \\
      
      \hline
      Training data  & 729 & 788 \\
      Testing data  & 186 & 193 \\
    \end{tabular}
  \end{center}
  \caption{SDCNL Dataset Statistics}
\end{table}

In this dataset, $729$ out of $1517$ instances are labelled as depression specific posts as shown in Table~2. The dataset is manually labelled to reduce noisy automated labels. The idea behind using this data is that we hypothesise that this dataset is even more complex than the Dreaddit dataset due to the presence of similar domain-specific words in posts. 

\subsection{Experimental Setup}
The original and the augmented dataset used for experimentation is quite noisy as the posts used in this data is user-generated natural language text expressing the feelings of the writer. The pre-processing steps are applied using the NLTK library\footnote{https://www.nltk.org/} of Python \cite{bird2006nltk}. The data is transformed before applying the supervised learning models employed in this work. The posts are long paragraphs, so in the first step the data is tokenized into sentences and then sentences are further tokenized into words. After removal of stop-words, punctuations,unknown characters from the extracted tokens, we use stemming and lemmatization to extract the root words.

After pre-processing of the data, it is transformed to a feature vector using Term Frequency- Inverse Document Frequency (TF-IDF), Word2Vec (W2V) \cite{goldberg2014word2vec} and Doc2Vec (D2V) \cite{lau2016empirical}. W2V embedding and D2V embedding provides dense vector representation of data while capturing its context. In this research work, the Gensim library\footnote{https://pypi.org/project/gensim/} is used to learn word embeddings from the training corpus using skip-gram algorithm. A vector of 300 dimensions is chosen and default settings of W2V and D2V models are used for experiments and evaluation.

The learning based classifiers which are used for this research work are the Logistic Regression (LR), the Support Vector Machine(SVM), and the Random Forest (RF) with the default settings of scikit-learn\footnote{https://scikit-learn.org/stable/} (sklearn) library of Python. The hardware configuration of the system which is used to perform this study is 2.6 GHz 6‑core Intel Core i7, Turbo Boost up to 4.5 GHz, with 12 MB shared L3 cache. 

\subsection{Experimental Results}
We reference \cite{kumar2021data} for implementation~\footnote{https://github.com/varunkumar-dev/TransformersDataAugmentation}  and use AugBERT, AugEDA, and AugBT on two datasets- Dreaddit and SDCNL. The dataset is divided into 75\% training and 25\% testing set and the value of Precision (P), Recall(R) and F1 score (F1) are computed on the testing samples to evaluate the performance of the classifiers with and without domain -specific data augmentation for mental health classification. Table~3 and Table~4 presents the results achieved for original and augmented data for Dreaddit and SDCNL using three different classifiers, namely, Logistic regression (LR), Support Vector Machine (SVM) and Random Forest (RF), respectively.

\begin{table*}[!h]
\begin{center}
\label{tab:table3}

\includegraphics[width=\textwidth]{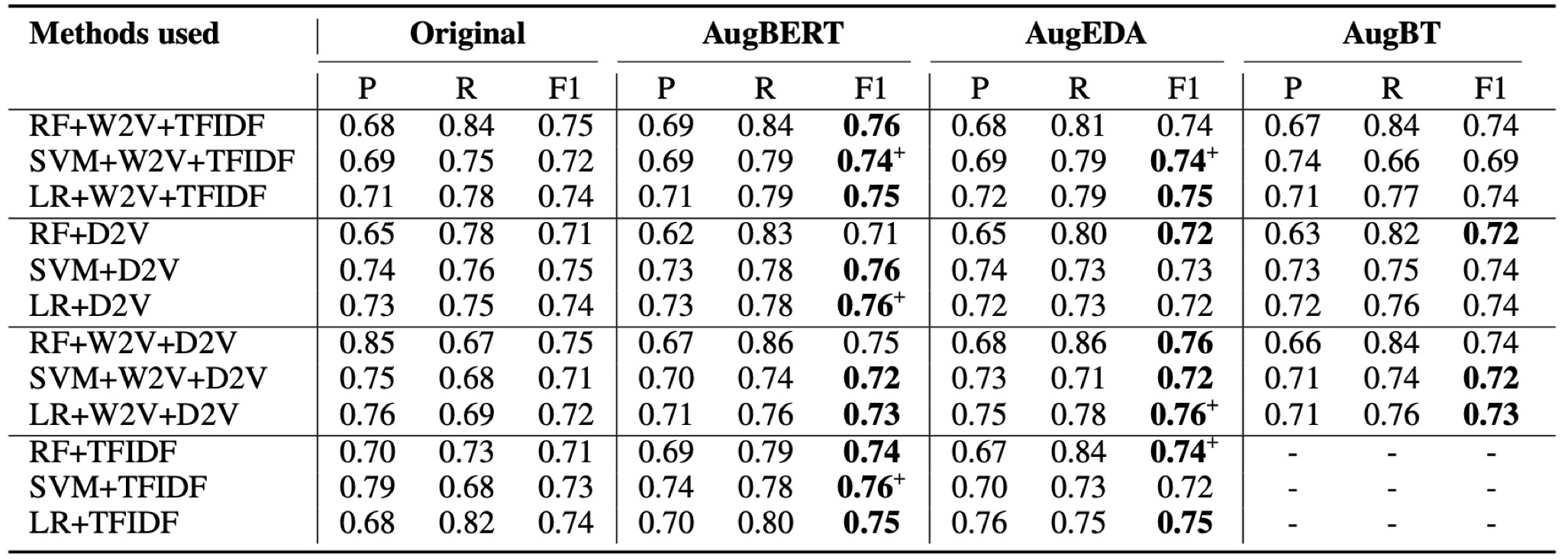}

\end{center}
\caption{Classification Results on Dreaddit Dataset: Precision(P), Recall(R), F-measure(F1) score on Original and Augmented Datasets using BERT, EDA and Back Translation. Text in bold shows the maximum F1 score achieved by the model. '-' indicates no results.'+' indicates significantly different results using statistical t-test.}
\end{table*}

\subsubsection{Experimental Results for Dreaddit}
As observed from Table~3, the F1 score showed an average improvement of around $1.4\%$ achieved by all models with AugBERT as compared to the original training dataset. It is also found that the AugEDA gives maximum improvement of around $4\%$ when W2V and D2V embeddings were employed with LR. Also, there is negligible improvement in the results with AugBT. 
 
\begin{table*}
\begin{center}

\label{tab:table4}
\includegraphics[width=\textwidth]{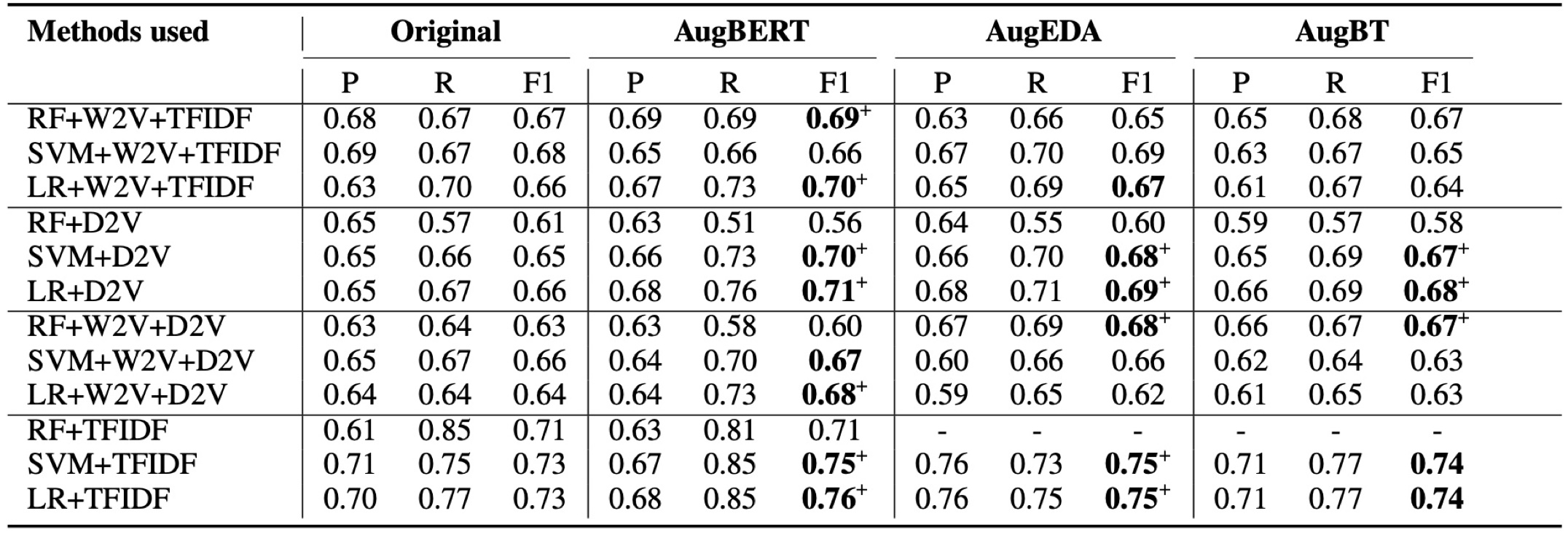}
\end{center}
\caption{Classification Results on SDCNL Dataset: Precision(P), Recall(R), F-measure(F1) score on Original and Augmented Datasets using BERT, EDA and Back Translation. Text in bold shows the maximum F1 score achieved by the model. '-' indicates no results.'+' indicates significantly different results using statistical t-test.}
\end{table*}

\subsubsection{Experimental Results for SDCNL}
In this Section, the results of the experimental study are presented for the SDCNL dataset. As observed from Table~4, the average improvement of around $2.3\%$ is observed for all the models as per F1 score with AugBERT. The AugEDA shows maximum improvement of more than $5\%$  when W2V and D2V embeddings were employed with RF. The results also indicate a minor improvement of around $1-2 \%$  when classifiers employed D2V and TF-IDF embeddings for representing augmented data using Back Translation. 

Due to increase in the size of augmented data, the input vector representations using TF-IDF requires higher computational time as compared to other embeddings. Thus, a few results are shown empty in Table~3 and Table~4. In healthcare, more \emph{precise} results are expected than \emph{recall} which means that the content which is identified as stressful must be correct and matters more than diagnosing the total number of correct instances. Thus, precision must improve more than recall values. We have considered these nuances to examine the results of classifiers and found that Logistic Regression gives improved results with the D2V encoding scheme.

\subsection{Data Diversity of Augmented Data}
The diversity of the generated data by different augmentation techniques are measured by the Bilingual Evaluation Understudy (BLEU) score \cite{papineni2002bleu}. The BLUE score ranges between $0$ and $1$. The lower the value, the better is the diversity in the data. Thus, the BLEU score is computed by comparing n-grams of both original and generated text where $n=2$.

\begin{table}[h!]
  \begin{center}

    \label{tab:table5}
    \begin{tabular}{l|c|c} 
      \textbf{ } & \textbf{Dreaddit} & \textbf{SDCNL} \\
      
      \hline
      \textbf{AugEDA}  & 0.97 & 0.99 \\
      \textbf{AugBERT}  & 0.82 & 0.97 \\
      \textbf{AugBT}  & 0.88 & 0.99 \\
    \end{tabular}
  \end{center}
      \caption{Data Diversity using BLEU Score}
\end{table}
As observed from Table~5, the BLEU score for augmented data varies from $82\%$ - $99\%$. The training samples are multiplied by $1.75$ to $2.0$ times for data augmentation approaches. The data for AugBERT is more diversified and thus, the results are significantly improved for AugBERT rather than AugEDA and AugBT as evident from Table~3 and Table~4. The experimental results show that the samples are upto $18\%$ more diverse than those of original training samples for AugBERT over the Dreaddit dataset. However, the least data diversity is observed for AugEDA and AugBT over the SDCNL dataset. 
\subsection{Statistical Significance}
In this Section, to understand the importance of generating more instances in training data is performed using three different data augmentation techniques. The statistical student's t-test was used to test the significance of the improvement in classifier using augmented data with $p-value$ as 0.05, 0.10, and 0.15. The resulting value for t-test in Dreaddit and SDCNL over AugBERT is obtained as 0.00033 and 0.09241 which shows the overall significant improvements with 5\% and 10\% significant levels, respectively. The results are improved in $83\%$, and $66\%$ in the cases of different encoding vectors and classifiers which are used as learning based algorithms for AugBERT and AugEDA data augmentation techniques, respectively. 

\subsubsection{Statistical Significance for Dreaddit}
It is evident from Table~6 that AugBERT and AugEDA show significantly improved results and there is no effect of AugBT over domain-specific data augmentation for mental health. 

\begin{table}[h!]
  \begin{center}
    \label{tab:table6}
    \begin{tabular}{l|c|c|c} 
        \textbf{Dreaddit} & \textbf{AugBERT} & \textbf{AugEDA} & \textbf{AugBT} \\
      
      \hline
      \textbf{t-test}  & -4.69041  & 1.07605 & 0.75593

 \\
        \textbf{p-value} &   0.00033     & 0.15247 & 0.23568 \\
        
        \hline
    
    \end{tabular}
  \end{center}
     \caption{Statistical Significance of overall results with Original Data}
 
\end{table}

On drilling down the results, it is observed that the AugBERT based augmented results for SVM classifier are significantly better than the other classification techniques. Some more significant improvements with the use of LR classifier is observed as shown in Table~3 with as high as 5\% for AugEDA. The variation of improvement in results ranges upto 4.1\%, 5.5\% and 1.3\% for AugBERT, AugEDA and AugBT, respectively.

\subsubsection{Statistical Significance for SDCNL}

The significant improvements over SDCNL dataset is observed on the basis of $p-value$ as $0.05$, $0.10$ and $0.15$ as shown in Table~7. The results have shown that the AugBERT and AugEDA gives better results for $10\%$ variation in results and validates the hypothesis that the augmented data gives significant improvements over the original dataset. 

\begin{table}[h!]
  \begin{center}
    \label{tab:table7}
    \begin{tabular}{l|c|c|c} 

      \textbf{SDCNL} & \textbf{AugBERT} & \textbf{AugEDA} & \textbf{AugBT} \\
      
      \hline
      \textbf{t-test}  & -1.42426  & -1.6361 & 0.25118

 \\
        \textbf{p-value} &  0.09241     & 0.06644 & 0.40338 \\

    \end{tabular}
  \end{center}
     \caption{Statistical Significance of overall results with Original Data}
 
\end{table}


Similar to the Dreaddit observations, the significant improvements with LR classifier are observed for classifying mental health into clinical depression and suicidal tendencies. On the contrary, SVM with D2V shows much better results with AugBERT, AugEDA and AugBT.

\section{Conclusion}
In this work, we use the data augmentation approach for mental health classification on two different social media datasets. The experimental results using Logistic Regression classifier and D2V embedding shows significant improvements in F1 score and Precision with AugBERT. To tackle the problem of data sparsity and support the automation of the 3-Step theory over social media data~\cite{klonsky2015three}, the data augmentation over mental healthcare may give remarkable results. In future, we are planning to use other domain-specific libraries and neural machine translation for explainable and conditional data augmentation.

\bibliography{acl2020.bbl}
\newpage
\appendix
\section {Appendix}
\begin{center}
    \label{appendix:table1}
  \includegraphics[width=\textwidth]{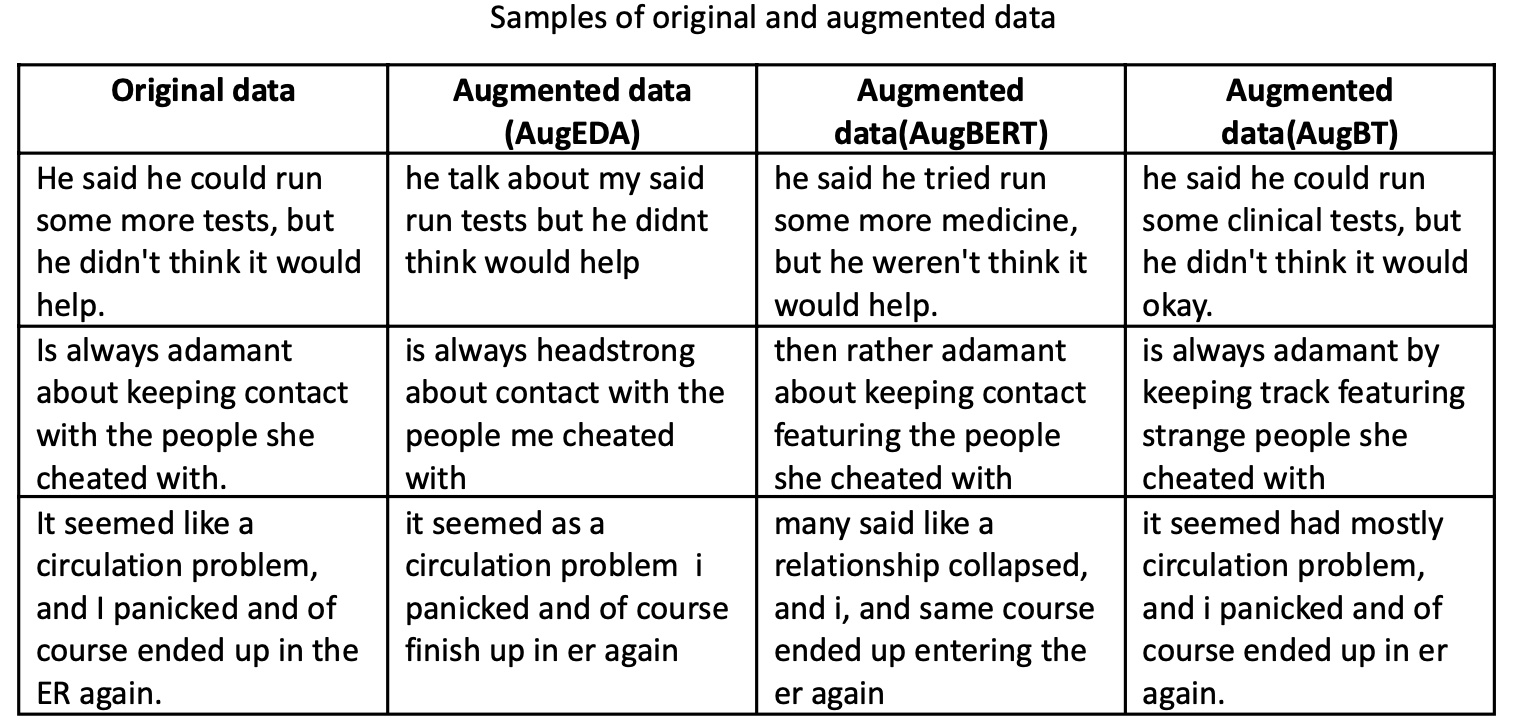}
\end{center}

\end{document}